%% file: sample-sigconf.tex
\documentclass[sigconf]{acmart}
\usepackage{hyperref}
\usepackage{enumitem}
\usepackage{float}
\usepackage{dblfloatfix}

\usepackage{placeins}
\usepackage{colortbl}
\usepackage{multirow} 
\usepackage[table,xcdraw]{xcolor}
\usepackage{tabularx}
\usepackage{bm}

\renewcommand{\arraystretch}{1.1} % 设置表格行间距的因子

\AtBeginDocument{%
  }

\input{8cmd}

\copyrightyear{2026}
\acmYear{2026}
\setcopyright{cc}
\setcctype{by}
\acmConference[KDD '26]{Proceedings of the 32nd ACM SIGKDD Conference on Knowledge Discovery and Data Mining V.2}{August 09--13, 2026}{Jeju Island, Republic of Korea}
\acmBooktitle{Proceedings of the 32nd ACM SIGKDD Conference on Knowledge Discovery and Data Mining V.2 (KDD '26), August 09--13, 2026, Jeju Island, Republic of Korea}
\acmDOI{10.1145/3770855.3817640}
\acmISBN{979-8-4007-2259-2/2026/08}

\begin{document}

\title{PHASE: Physiology-Aware Hyperspectral Reconstruction via Object-to-Human Domain Adaptation}

\author{Yufei Wen}
\affiliation{%
  \institution{The Hong Kong University of Science and Technology (Guangzhou)}
  \city{Guangzhou}
  \state{Guangdong}
  \country{China}
}
\email{ywen196@connect.hkust-gz.edu.cn}

\author{Shuxin Zhong}
\affiliation{%
  \institution{The Hong Kong University of Science and Technology (Guangzhou)}
  \city{Guangzhou}
  \state{Guangdong}
  \country{China}
}
\email{shuxinzhong@hkust-gz.edu.cn}

\author{Jingdan Kang}
\affiliation{%
  \institution{South China University of Technology}
  \city{Guangzhou}
  \state{Guangdong}
  \country{China}
}
\email{jingdankang6@gmail.com}

\author{Yuting Zhang}
\affiliation{%
  \institution{The Hong Kong University of Science and Technology (Guangzhou)}
  \city{Guangzhou}
  \state{Guangdong}
  \country{China}
}
\email{yzhang430@connect.hkust-gz.edu.cn}

\author{Jintai Chen}
\affiliation{%
  \institution{The Hong Kong University of Science and Technology (Guangzhou)}
  \city{Guangzhou}
  \state{Guangdong}
  \country{China}
}
\email{jintaichen@hkust-gz.edu.cn}

\author{Kaishun Wu}
\authornote{Corresponding author.}
\affiliation{%
  \institution{The Hong Kong University of Science and Technology (Guangzhou)}
  \city{Guangzhou}
  \state{Guangdong}
  \country{China}
}
\email{wuks@hkust-gz.edu.cn}

\renewcommand{\shortauthors}{Y. Wen et al.}

\input{0abstract}




\maketitle
\newcommand\kddavailabilityurl{https://doi.org/10.5281/zenodo.20305653}
\newcommand\kddrepourl{https://github.com/Dreamer1209/PHASEKDD}
\ifdefempty{\kddavailabilityurl}{}{
\begingroup\small\noindent\raggedright\textbf{Resource Availability:}\\
The source code of this paper has been made publicly available at \\
\href{\kddavailabilityurl}{https://doi.org/10.5281/zenodo.20305653}, with the development repository at \href{\kddrepourl}{https://github.com/Dreamer1209/PHASEKDD}.
\endgroup
}
\input{1intro}
\input{2relatedwork}

\input{4framework}
\input{5experiments}
\input{7conclusion}

% \newpage

%%
\begin{acks}
This work was supported in part by the China NSFC Grant\\(No.~62472366), the Project of the Department of Education of Guangdong Province (Nos.~2023KCXTD042, 2024GCZX003, \\2024KCXTD008, and 2025KCXTD056), the Guangdong Provincial Key Laboratory of Integrated Communication, Sensing and Computation for Ubiquitous Internet of Things (No.~2023B1212010007), the `111 Center' (No.~D25008), the Shenzhen Science and Technology Foundation (No.~ZDSYS20190902092853047), the Guangdong Basic and Applied Basic Research Foundation (No.~2026A1515011793), and the Youth S\&T Talent Support Programme of Guangdong Provincial Association for Science and Technology (No.~SKXRC2025467).
\end{acks}

\newpage
\bibliographystyle{ACM-Reference-Format}
\bibliography{9bib}

\end{document}

%% file: 8cmd.tex
\newcommand{\N}{\texttt{PHASE}}

\newcommand{\ComponentA}{Physiological Channel Reinterpretation}
\newcommand{\ComponentB}{Physiologically Constrained Alignment}

%% file: 0abstract.tex
\begin{abstract}
Although hyperspectral imaging offers unparalleled non-invasive physiological insight,
its bulky hardware, slow acquisition, and regulatory burden severely limit its clinical availability.
A natural workaround is to reconstruct hyperspectral information from ubiquitous RGB or CASSI measurements.
However, existing paradigms, developed for object-centric scenes, rely on reflectance-based feature alignment, assuming that spectral similarity preserves semantic meaning.
This assumption breaks down in physiological imaging, where visually similar RGB responses may arise from distinct and entangled physiological states.
This mismatch motivates a shift from reflectance alignment to physiology-aware representation learning, grounded in shared light–matter interaction principles—a shift that introduces fundamental challenges from cross-channel semantic shifts (C1) and irreversible information loss in RGB-based acquisition (C2).
We therefore design \N, a physiology-aware hyperspectral reconstruction paradigm that fundamentally redefines object-to-human transfer by disentangling cross-channel physiological semantics via \ComponentA and restricting reconstruction to physiologically plausible solutions through \ComponentB.
Under two source-to-target transfer protocols, \N\ consistently outperforms state-of-the-art methods by up to $+$2.20 SSIM and $-$3.06 in SAM with merely 1.5\% labeled supervision.

\end{abstract}

%% file: 1intro.tex
\section{Introduction}
\label{sec:intro}

Hyperspectral imaging (HSI) captures contiguous spectral signatures that encode rich physical and biochemical properties~\cite{khan2018modern, peyghambari2021hyperspectral, nisha2022current}.
This capability enables non-invasive physiological characterization and downstream tasks such as cancer margin detection and hemoglobin estimation~\cite{barberio2021intraoperative, leon2020non, aboughaleb2020hyperspectral, skjelvareid2017detection, sicher2018hyperspectral}.
However, the bulky hardware, slow acquisition, and stringent ethical constraints of HSI severely limit its deployment in routine clinical practice~\cite{ng2024hyper}, 
creating a persistent mismatch between its diagnostic potential and real-world availability.
This tension motivates a fundamental question:
\textit{Can we acquire hyperspectral information from ubiquitous, low-cost measurements?}

Recent efforts~\cite{zhang2025cross, wang2022semi, fang2022semi} address this question by reconstructing HSI from RGB or Coded Aperture Snapshot Spectral Imaging (CASSI) measurements~\cite{zhang2023essaformer, brorsson2021reconstruction, zhang2011reconstructing} using neural networks, 
ranging from convolutional models that capture local spectral–spatial correlations~\cite{li2017hyperspectral, wang2019hyperspectral, chang2018hsi} to Transformer-based architectures that model global dependencies~\cite{cai2022mask, cai2022coarse}.
While these methods achieve strong performance on object-centric datasets~\cite{chakrabarti2011statistics, yasuma2010generalized, arad2022ntire, bandara2021hyperspectral}, 
their generalization to medical imaging scenarios remains limited~\cite{guan2021domain, wilm2023mind}.

We argue that this limitation stems from a fundamental mismatch in modeling assumptions rather than data scarcity or model capacity. 
\textit{In object-centric scenes, spectra are dominated by surface reflectance, under which spectral similarity reliably preserves semantic meaning~\cite{adelson2001seeing, dror2001statistics}. 
In contrast, spectra from human tissue arise from volumetric absorption, scattering, and physiological composition~\cite{jacques2013optical, tuchin2015tissue, wang2007biomedical},
such that visually similar measurements may correspond to distinct underlying biological states}.

Consequently, directly performing cross-domain reflectance alignment can encourage shortcut learning on superficial color and reflectance cues rather than the underlying physiological mechanisms, limiting generalization in medical imaging.
Building on \textbf{the shared light–matter interaction principles}, we therefore reposition object-to-human domain adaptation for hyperspectral reconstruction from a reflectance alignment problem to one of physiology-aware representation learning.

\begin{figure}[t!]
\centering
\includegraphics[width=1\columnwidth]{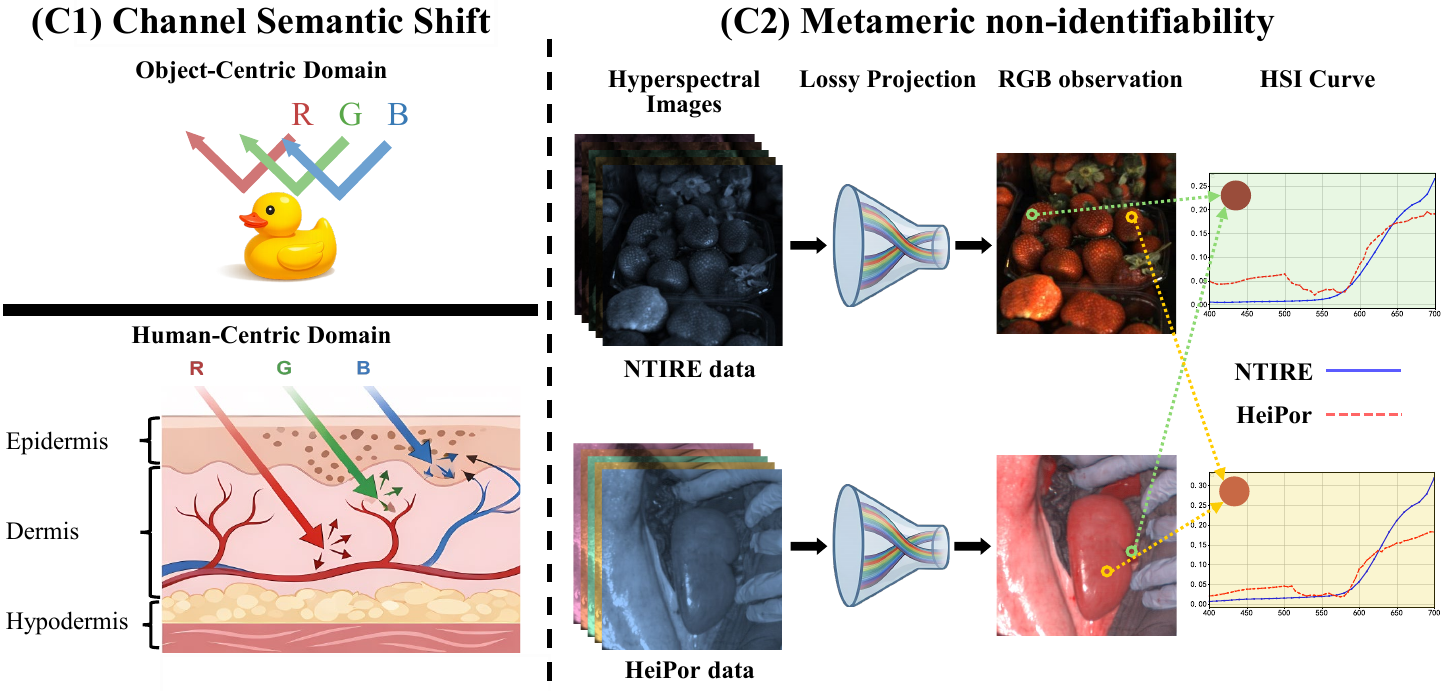}
\caption{
Illustration of the domain adaptation challenges. (C1) Channel Semantic Shift: RGB channels encode surface reflectance in object-centric scenes versus tissue absorption in human imaging. (C2) Metameric non-identifiability: The lossy RGB projection creates spectral ambiguity, making the inverse problem structurally underdetermined.
}
\Description{A two-panel schematic illustrating cross-channel semantic shift and metameric ambiguity in object-to-human hyperspectral reconstruction.}
\label{teaser}
\end{figure}

Despite this shared physical principle, object-to-human domain adaptation for hyperspectral reconstruction still faces two fundamental challenges, as illustrated in Figure~\ref{teaser}:
\begin{itemize} [leftmargin=*]
    \item \textbf{(C1) Physiology-induced channel semantic shift.} 
    RGB channels carry fundamentally different semantics across domains~\cite{sun2016deep, peng2019moment}.
    In object-centric scenes, they primarily encode surface reflectance and material color~\cite{geusebroek2001color}, 
    whereas in human tissue each channel is dominated by distinct physiological absorption and scattering processes~\cite{anderson1981optics, zonios2001skin}.

    \item \textbf{(C2) Metameric non-identifiability under lossy observation.}
    RGB measurements act as a severely compressed and lossy projection of the underlying physiological state space~\cite{tsumura2003image, claridge2003colour}.
    Distinct hyperspectral signatures can therefore correspond to indistinguishable RGB observations~\cite{wyszecki2000color}, 
    inducing an intrinsic many-to-one mapping that renders RGB-to-hyperspectral reconstruction fundamentally underdetermined.
\end{itemize}

To this end, we propose~\N, a \underline{PH}ysiology-\underline{A}ware hyper\underline{S}pectral r\underline{E}construction paradigm built upon Mean Teacher~\cite{tarvainen2017mean} that explicitly grounds object-to-human transfer in physiological semantics.
We introduce two complementary components.
To address \textbf{C1}, \textit{\ComponentA} applies spectral-complexity--guided, channel-adaptive masking to the student branch, suppressing spurious channel shortcuts and encouraging physiologically consistent cross-channel reasoning. 
To tackle \textbf{C2}, \textit{\ComponentB} regularizes the underdetermined RGB-to-HSI mapping via reliability-gated prototype distribution alignment between teacher and student representations, preventing source-biased pseudo-supervision from dominating target adaptation. 
Together, these components yield more stable self-training and stronger cross-domain generalization in clinical scenarios.
Our contributions are summarized as follows:
\begin{itemize} [leftmargin=*]
    \item 
    We uncover a previously overlooked assumption failure in object-to-human domain adaptation for hyperspectral reconstruction: \textit{reflectance-based alignment preserves semantics for objects but fundamentally breaks down for physiological imaging}.
    To resolve this, we reframe reconstruction as \textit{physiology-aware representation learning grounded in shared light–matter interaction principles}.

    \item 
    We implement \N\ with two key designs: (i) \textit{\ComponentA} reconfigures channel semantics by enforcing cross-channel spectral inference under physiological constraints; 
    and (ii) \textit{\ComponentB} constrains non-identifiable RGB-to-hyperspectral reconstruction by suppressing physiologically implausible solutions and promoting clinically grounded ones.

    \item
    On NTIRE2020-to-Hyper-Skin and NTIRE2022-to-Hyper-Skin adaptation, 
    \N\ achieves +1.01 and +2.20 SSIM improvements and $-1.17$ and $-3.06$ SAM gains over the strongest baselines using only 1.5\% labeled target data.
\end{itemize}

%% file: 2relatedwork.tex
\section{Related Work}
\label{sec:related_work}

\subsection{Hyperspectral Image Reconstruction}
Hyperspectral image (HSI) reconstruction aims to recover high-dimensional spectral information from low-dimensional or compressed observations~\cite{ghamisi2018advances, arad2022ntire}. 
Existing approaches can be broadly categorized into two major directions based on the input modality: RGB-based~\cite{arad2016sparse, shi2018hscnn, cai2022mst++} and CASSI–based methods~\cite{wagadarikar2008single, miao2019net, meng2020end}.

RGB-based reconstruction~\cite{yan2020reconstruction,zhang2023essaformer,zhao2023hsgan} aims to reconstruct HSI from RGB images by exploiting statistical correlations between RGB measurements and hyperspectral signatures. 
Early works employ CNNs to model local spectral–spatial correlations~\cite{yan2020reconstruction}, while recent approaches adopt Transformer architectures to capture long-range spectral dependencies and global context~\cite{zhang2023essaformer,wang2025hypersigma}. 

CASSI-based reconstruction recovers HSI from compressed measurements obtained via optical modulation~\cite{cai2022mask, chen2024hyperspectral, chen2025low, chen2023spectral}. 
Due to the severe ill-posedness of the inverse problem, 
existing approaches typically exploit the inherent redundancy and compressibility of hyperspectral data~\cite{xue2021spatial, zhang2021learning, cai2022coarse}. 
Representative strategies include masking-based regularization~\cite{cai2022mask} and unrolled optimization frameworks~\cite{chen2024hyperspectral}, which explicitly embed physical priors or iterative solvers to improve reconstruction stability and convergence.

\textbf{Summaries.}
Despite their progress, most RGB-to-HSI and CASSI reconstruction models rely on \textbf{large-scale paired training data and consistent acquisition conditions}, 
both of which are difficult to obtain in medical settings due to high annotation cost and strict privacy or ethical constraints.

\subsection{Semi-Supervised Domain Adaptation}
Semi-supervised domain adaptation (SSDA) has achieved notable success in mitigating domain shift and label scarcity in vision tasks~\cite{yang2022survey,van2020survey}. 
Existing methods fall into three paradigms: 
\begin{itemize} [leftmargin=*]
    \item Consistency-regularized pseudo-labeling methods generate pseudo-labels for unlabeled target samples and enforce consistency under perturbations~\cite{feng2023pseudo}.
    Representative methods include AdaMatch~\cite{berthelot2021adamatch}, which employs adaptive confidence thresholds to select reliable pseudo-labels, 
    and MCL-SDA~\cite{yan2022multi}, which introduces multi-view consistency across classifiers to stabilize training.
    \item Mixup-based approaches reduce domain discrepancy by interpolating source and target samples~\cite{zhou2022context}. 
    IDM-SDA~\cite{li2024inter} applies inter-domain mixup at the input level, while SFPM-SDA~\cite{ma2023source} progressively mixes labeled and pseudo-labeled data for robustness.
    \item Prototype-based methods align source and target distributions using representative prototypes. 
    PRO-SDA~\cite{huang2023semi} learns class-agnostic prototypes, 
    while Unmix-SDA~\cite{baghbaderani2023unsupervised} derives domain-specific endmembers via spectral unmixing.
\end{itemize}

\textbf{Summaries.}
Despite their success in natural images, existing SSDA methods presuppose shared semantics and compatible feature formation across domains. 
Across the object–human divide, \textbf{channel semantics are redefined and spectral generation is fundamentally altered}, breaking these assumptions at their core.

%% file: 4framework.tex
\begin{figure*}[t!]
\centering
\includegraphics[width=1\textwidth]{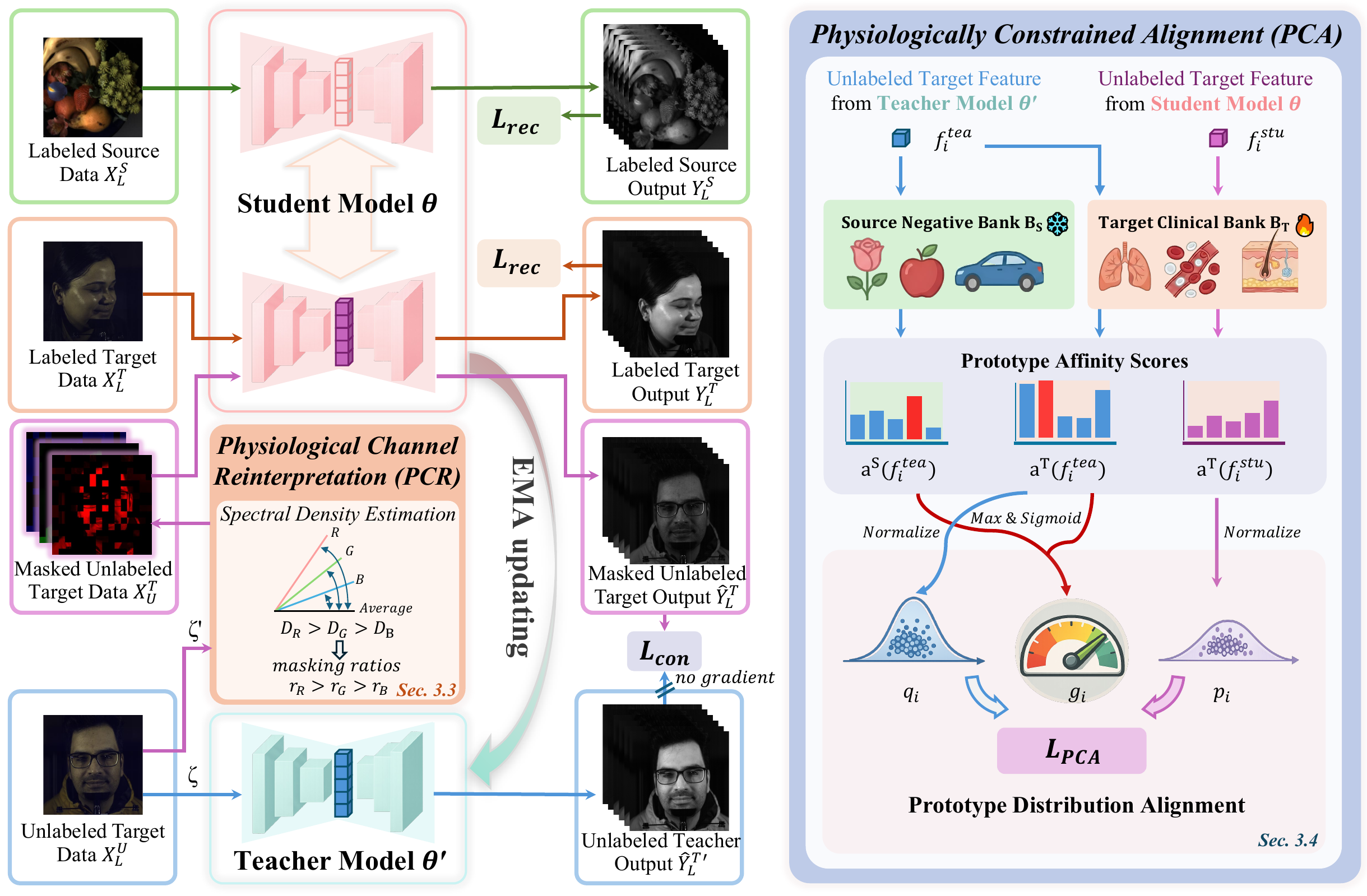}
\caption{
Overview of \N, an object-to-human hyperspectral reconstruction framework built on the Mean Teacher~\cite{tarvainen2017mean}.
Given the same input, a strong augmentation $\zeta'$ and a weak augmentation $\zeta$ are applied to the student and teacher networks, respectively.
The student and teacher share identical architectures, with the teacher parameters $\theta'$ updated as the exponential moving average (EMA) of the student parameters $\theta$. 
On top of this backbone, \N\ introduces two domain-specific regularization modules.
PCR enforces cross-channel spectral consistency through adaptive masking, while
PCA promotes physiological plausibility by aligning representations to reliable prototypes via confidence-gated consistency.
}
\Description{An overview of the PHASE architecture: a Mean Teacher backbone augmented with the PCR and PCA regularization modules.}
\label{framework}
\end{figure*}

\section{Methodology}
\label{sec:method}

\subsection{Problem Formulation}
Hyperspectral image (HSI) reconstruction aims to recover a high-dimensional spectral image 
$y \in \mathbb{R}^{H \times W \times C}$ from a conventional RGB input 
$x \in \mathbb{R}^{H \times W \times 3}$, where $C$ denotes the number of spectral bands. 
In this work, we study \textit{object-to-human hyperspectral reconstruction} in clinical settings, 
which transfers spectral knowledge learned from object-centric scenes to human tissue under extremely limited supervision.
Formally, we are given a labeled source dataset 
$\mathcal{D}_{l}^{S} = \{(x_i^S, y_i^S)\}_{i=1}^{L_S}$ collected from general object-centric scenes, 
and a human-centered target dataset consisting of a small labeled subset 
$\mathcal{D}_{l}^{T} = \{(x_i^T, y_i^T)\}_{i=1}^{L_T}$ and a large unlabeled subset 
$\mathcal{D}_{u}^{T} = \{x_i^T\}_{i=1}^{U_T}$, where $L_T \ll L_S$ and $L_T \ll U_T$. 
Our goal is to learn a reconstruction model $f_\theta$ that generalizes to unseen human tissue 
by leveraging labeled object-centric data and a small set of labeled target samples.

\subsection{Framework Overview}
Figure~\ref{framework} provides an overview of~\N, a physiology-aware hyperspectral reconstruction paradigm for object-to-human transfer.
Methodologically, \N\ is instantiated within a semi-supervised domain adaptation (SSDA) framework based on the Mean Teacher paradigm~\cite{tarvainen2017mean}, where a student network is optimized with supervised and consistency objectives, and a teacher network is updated via exponential moving average (EMA).
Crucially, unlike conventional SSDA methods that treat domain shift as a feature distribution mismatch, 
\N\ targets the more fundamental discrepancy in object-to-human reconstruction—namely, \textit{channel semantic mismatch and structural non-identifiability in spectral reconstruction}—by introducing two complementary components.
\begin{itemize}[leftmargin=*]

    \item \textit{\ComponentA} is a spectral--complexity-–guided channel--adaptive masking strategy designed to address channel semantic mismatch.
    By selectively suppressing spectrally dominant channels, it disrupts source-domain channel shortcuts and enforces cross-channel reasoning, enabling the model to capture channel dependencies consistent with physiological absorption and scattering processes.

    \item \textit{\ComponentB} explicitly suppresses physiologically invalid yet RGB-consistent explanations while biasing reconstruction toward clinically grounded spectra.
    Specifically, it employs a source negative bank to repel source-domain non-physiological patterns and a target clinical bank to anchor clinical prototypes. By contrasting affinities to these banks, we derive a reliability gate that modulates the alignment of target-bank prototype distributions, steering adaptation toward physiologically plausible representations.
    
\end{itemize}

\subsection{\ComponentA}
\label{subsec:component}
\begin{figure}[htbp]
\includegraphics[width=1\columnwidth]{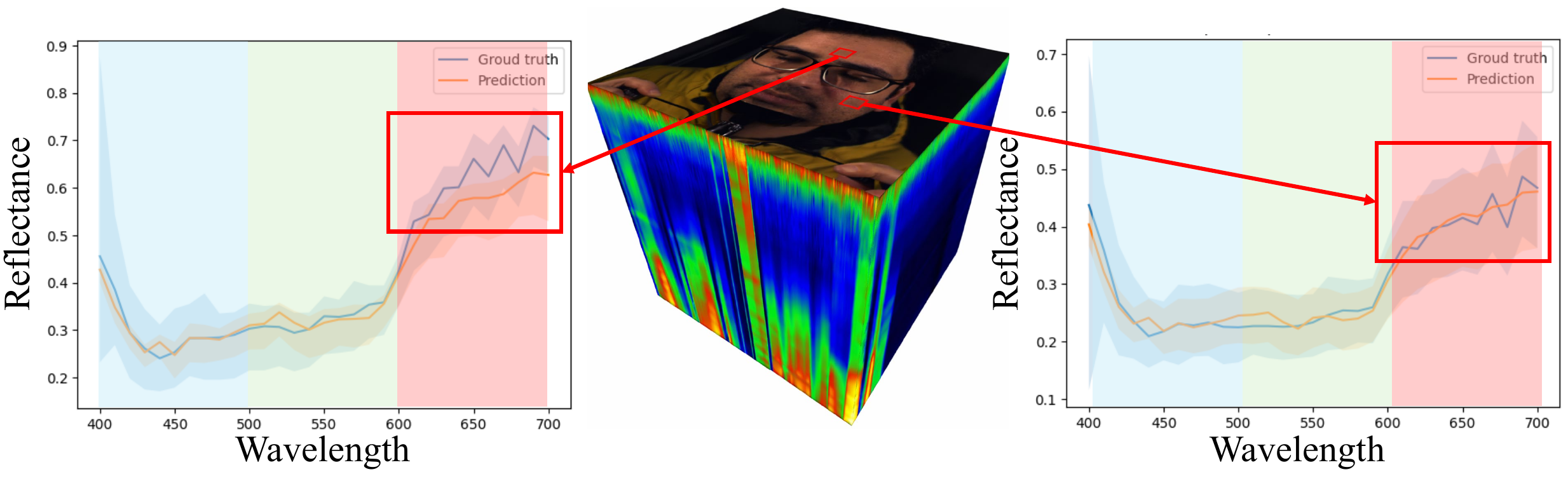}
\caption{
Skin exhibits substantial spectral density variation, characterized by pronounced complexity in the red band.
}
\Description{Reflectance-versus-wavelength plots of facial skin showing higher spectral variability in the red band.}
\label{humanred}
\end{figure}

\noindent\textbf{Motivation.}
Masking is widely used in reconstruction and representation learning to promote reasoning over incomplete inputs and to enforce consistency under perturbations~\cite{tarvainen2017mean, zhang2024maskfusionnet, bae2016dct, chen2019multi}.
However, most existing masking strategies are channel-agnostic and implicitly assume that spectral information is evenly distributed across RGB channels~\cite{he2022masked, xie2022simmim}. 
This assumption breaks down in human tissue imaging, 
where absorption–scattering mechanisms induce highly uneven spectral contributions across wavelengths~\cite{jacques2013optical}. 
As shown in Figure~\ref{humanred}, facial skin exhibits relatively stable reflectance in the blue–green wavelength region, while the red wavelength region contains substantially richer and more variable spectral structures.
As RGB channels integrate hyperspectral bands over distinct wavelength regions, the red-region complexity is funneled into the R channel rather than being evenly shared.
Consequently, object-centric models often over-emphasize spectrally dominant channels, leading to domain-specific shortcuts that fail to transfer to human tissue.
Thus, we design \textit{\ComponentA} (PCR), a spectral-complexity–guided channel-adaptive masking strategy that preferentially masks spectrally dominant channels to suppress domain-specific shortcuts and enforce cross-channel spectral reasoning under severe semantic drift.

\noindent\textbf{Spectral Perturbation.}
Given an RGB image $\mathbf{x}\in\mathbb{R}^{H\times W\times 3}$, we reshape it into a pixel-wise matrix $\tilde{\mathbf{x}}\in\mathbb{R}^{N\times 3}$ with $N=H\cdot W$.
For each channel $b\in\{\mathrm{R},\mathrm{G},\mathrm{B\}}$, we construct a perturbed matrix $\tilde{\mathbf{x}}^{(b)}$ by replacing channel $b$ with its spatial mean:
\begin{equation}
\tilde{\mathbf{x}}^{(b)}=\frac{1}{N}\sum_{i=1}^{N}\tilde{\mathbf{x}}_{i,b}.
\end{equation}
\textbf{Spectral Density Estimation.}
We quantify the spectral density $\mathcal{D}_b$ of channel $b$ as the Spectral Angle Mapper (SAM) deviation between $\tilde{\mathbf{x}}$ and $\tilde{\mathbf{x}}^{(b)}$, which serves as a proxy for the contribution of $b$ to preserving spectral structure~\cite{kruse1993spectral}:
\begin{equation}
\mathcal{D}_b=\text{SAM}(\tilde{\mathbf{x}}^{(b)},\tilde{\mathbf{x}}),
\end{equation}
where SAM captures directional changes in spectral signatures and is insensitive to magnitude variations.
A larger $\mathcal{D}_b$ indicates channel $b$ plays a more critical role in preserving the spectral structure.

\noindent\textbf{Adaptive Channel Masking.}
Masking dominant channels suppresses shortcut reliance and enforces cross-channel inference.
Based on the spectral densities $\{\mathcal{D}_b\}$, we assign higher masking ratios to channels with larger $\mathcal{D}_b$.
Specifically, channel-wise masking ratios $\mathbf{r}=[r_{\text{R}},r_{\text{G}},r_{\text{B}}]$ are obtained via min–max normalization:
\begin{equation}
r_b=r_{\min}+\frac{\mathcal{D}_b-\min(\mathcal{D})}{\max(\mathcal{D})-\min(\mathcal{D})}(r_{\max}-r_{\min}),
\end{equation}
where $\mathcal{D}=[\mathcal{D}_{\text{R}},\mathcal{D}_{\text{G}},\mathcal{D}_{\text{B}}]$.
For each channel $b$, a fraction $r_b$ of spatial blocks is randomly masked on the corresponding channels, forcing the model to reconstruct missing information from complementary channels and to capture cross-channel dependencies.

\subsection{\ComponentB}
\label{subsec:componentB}

\textbf{Motivation.}
Mean Teacher exploits unlabeled target data by enforcing prediction consistency between student and EMA teacher~\cite{tarvainen2017mean}.
However, under severe domain shift, \textit{metamerism} can make distinct spectra RGB-indistinguishable~\cite{stiles1962counting,fu2025limitations}, yielding biased pseudo-labels and negative transfer~\cite{wang2019characterizing}.
This risk is amplified by open-set regions (e.g., surgical backgrounds) that overlap with tissue in RGB space~\cite{allan20192017}.
To this end, we introduce \textit{\ComponentB} (PCA), which estimates teacher reliability via dual-bank affinities and applies reliability-gated distribution alignment to guide self-training toward physiologically plausible solutions.
This formulation parallels the Kubelka--Munk decomposition~\cite{anderson1981optics}: $\mathcal{B}_T$ provides data-driven tissue basis spectra, $\mathcal{B}_S$ encodes a reflectance prior, and the reliability gate (Eq.~\ref{eq:gate}) performs implicit spectral unmixing without requiring tissue-specific calibration.

We first define the representation space used for reliability estimation.
For an RGB input $x \in \mathbb{R}^{H \times W \times 3}$, the student and EMA-teacher networks produce intermediate feature maps $\mathbf{F}^{\mathrm{stu}}, \mathbf{F}^{\mathrm{tea}} \in \mathbb{R}^{H' \times W' \times C'}$.
We flatten the spatial grid and index each location by $i \in \{1, \ldots, H'W'\}$, yielding feature vectors $\mathbf{f}^{\mathrm{stu}}_{i}, \mathbf{f}^{\mathrm{tea}}_{i} \in \mathbb{R}^{C'}$.

\noindent\textbf{Dual-anchor Semantic Banks.}
We maintain two prototype banks in the EMA-teacher feature space as semantic anchors for reliability estimation.
The \textit{Source Negative Bank} $\mathcal{B}_S=\{\mathbf{b}^S_m\}_{m=1}^{M}$ is constructed by clustering teacher features from the labeled source set $\mathcal{D}_l^S$ and remains frozen, serving as a stable source-domain reference.
The \textit{Target Clinical Bank} $\mathcal{B}_T=\{\mathbf{b}^T_k\}_{k=1}^{K}$ is initialized from the scarce labeled target set $\mathcal{D}_l^T$ to capture clinically meaningful target modes and is updated online to track target-domain evolution:
\begin{equation}
\mathbf{b}^T_k \leftarrow \mu\,\mathbf{b}^T_k + (1-\mu)\,\bar{\mathbf{f}}^{\mathrm{tea}}_k,
\end{equation}
where $\bar{\mathbf{f}}^{\mathrm{tea}}_k$ denotes the mean of teacher features assigned to prototype $\mathbf{b}^T_k$ via nearest-prototype cosine similarity, and $\mu\in[0,1)$ is the momentum coefficient.

\noindent\textbf{Prototype Affinity Scores.}
For each feature vector $\mathbf{f}_i\in\mathbb{R}^{C'}$ at location $i$, we measure its similarity to the target and source prototype banks.
These affinity vectors provide the evidence used for reliability gating and also serve as logits for prototype-level distribution alignment.
Specifically, the affinities between $\mathbf{f}_i$ and the target bank $\mathcal{B}_T=\{\mathbf{b}^T_k\}_{k=1}^{K}$ are defined as:
\begin{equation}
\mathbf{a}^{T}(\mathbf{f}_{i})
=
\big[\mathrm{sim}(\mathbf{f}_i,\mathbf{b}^T_1),\ldots,\mathrm{sim}(\mathbf{f}_i,\mathbf{b}^T_K)\big]
\in\mathbb{R}^{K},
\end{equation}
while its affinities to the source reference bank $\mathcal{B}_S=\{\mathbf{b}^S_m\}_{m=1}^{M}$ are
\begin{equation}
\mathbf{a}^{S}(\mathbf{f}_{i})
=
\big[\mathrm{sim}(\mathbf{f}_i,\mathbf{b}^S_1),\ldots,\mathrm{sim}(\mathbf{f}_i,\mathbf{b}^S_M)\big]
\in\mathbb{R}^{M}.
\end{equation}
Here $\mathrm{sim}(\cdot,\cdot)$ denotes cosine similarity.

\noindent\textbf{Physiological Reliability Gating.}
For each EMA-teacher feature $\mathbf{f}^{\mathrm{tea}}_i\in\mathbb{R}^{C'}$, 
we summarize its alignment with the target and source banks by their strongest affinities:
\begin{equation}
\mathcal{A}_T(\mathbf{f}^{\mathrm{tea}}_i)=\max_{k}\ \mathbf{a}^{T}(\mathbf{f}^{\mathrm{tea}}_i)(k),\quad
\mathcal{A}_S(\mathbf{f}^{\mathrm{tea}}_i)=\max_{m}\ \mathbf{a}^{S}(\mathbf{f}^{\mathrm{tea}}_i)(m).
\end{equation}
Intuitively, a teacher feature is considered more \emph{physiologically plausible} if it is well explained by the target clinical prototypes (high $\mathcal{A}_T$) while remaining weakly aligned with source patterns (low $\mathcal{A}_S$).
We therefore define a reliability gate based on their affinity contrast:
\begin{equation}
g_i=\sigma\!\left(\frac{\mathcal{A}_T(\mathbf{f}^{\mathrm{tea}}_i)-\mathcal{A}_S(\mathbf{f}^{\mathrm{tea}}_i)}{\tau_g}\right)\in(0,1),
\label{eq:gate}
\end{equation}
where $\sigma(\cdot)$ denotes the sigmoid and $\tau_g$ is a temperature. 
The gate $g_i$ upweights locations whose teacher features are more target-consistent and suppresses source-like features, thereby reducing the influence of unreliable pseudo-signals during adaptation.

\noindent\textbf{Prototype Distribution Alignment.}
For unlabeled target samples $x_u^T\in\mathcal{D}_u^T$,
we align student and EMA-teacher representations in the clinical prototype space.
At each location $i$, target-bank affinities are normalized into prototype distributions.
Specifically, the teacher feature $\mathbf{f}^{\mathrm{tea}}_i$ and student feature $\mathbf{f}^{\mathrm{stu}}_i$ yield: 

\begin{equation}
\begin{aligned}
\mathbf{q}_i
&=
\mathrm{softmax}\!\left(\mathbf{a}^{T}(\mathbf{f}^{\mathrm{tea}}_i)/\tau_p\right)
\in \mathbb{R}^{K}, \\
\mathbf{p}_i
&=
\mathrm{softmax}\!\left(\mathbf{a}^{T}(\mathbf{f}^{\mathrm{stu}}_i)/\tau_p\right)
\in \mathbb{R}^{K},
\end{aligned}
\end{equation}
where $\mathbf{q}_i,\mathbf{p}_i\in\mathbb{R}^{K}$ and $\tau_p{=}0.2$ is the softmax temperature~\cite{hinton2015distilling}.
We then align the student distribution $\mathbf{p}_i$ to the teacher distribution $\mathbf{q}_i$ using a reliability-weighted soft cross-entropy:
\begin{equation}
\mathcal{L}_{\mathrm{PCA}}
=
-\frac{1}{H'W'}\sum_{i=1}^{H'W'} g_i \sum_{k=1}^{K} \mathbf{q}_i(k)\log \mathbf{p}_i(k),
\end{equation}
where the gate $g_i$ (Eq.~\ref{eq:gate}) emphasizes locations with reliable teacher signals while suppressing unreliable ones.

\subsection{The Optimization Process}

As shown in Figure~\ref{framework}, our overall objective promotes accurate reconstruction, cross-domain consistency, and physically plausible spectral structure.
The overall objective combines supervised reconstruction on labeled data and unsupervised regularization on unlabeled target data:
\begin{equation}
\mathcal{L} = \mathcal{L}_{\text{sup}} + \lambda_{\text{un}}(t) \cdot \mathcal{L}_{\text{un}},
\end{equation}
where $\mathcal{L}_{\text{sup}}$ enforces pixel-wise accuracy on all available labeled data, while $\mathcal{L}_{\text{un}}$ regularizes unlabeled target predictions via student-teacher consistency and prototype alignment.
To mitigate early training instability from unreliable pseudo-labels, $\lambda_{\text{un}}(t)$ follows a Gaussian ramp-up schedule, scaling the weight from 0 to $\lambda_{max}$.

\subsubsection{\texorpdfstring{Supervised Loss \(\mathcal{L}_{\text{sup}}\)}{Supervised Loss Lsup}}
The supervised objective is driven by a reconstruction loss $\mathcal{L}_{\text{rec}}$, which quantifies the pixel-wise discrepancy between the predicted and ground-truth spectral signatures.
To enforce high fidelity while maintaining robustness against spectral outliers, we formulate $\mathcal{L}_{\text{rec}}$ using the $\ell_1$ distance:

\begin{equation}
\mathcal{L}_{\text{rec}}(\hat{y}, y) = \mathcal{L}_{1}(\hat{y}, y).
\end{equation}
The supervised objective is applied to both labeled source and labeled target samples, as illustrated in Figure~\ref{framework}:
\begin{equation}
\mathcal{L}_{\text{sup}} = \mathcal{L}_{\text{rec}}(\hat{y}^{S}_L, y^{S}_L) + \mathcal{L}_{\text{rec}}(\hat{y}^{T}_L, y^{T}_L).
\end{equation}

\subsubsection{\texorpdfstring{Unsupervised Loss \(\mathcal{L}_{\text{un}}\)}{Unsupervised Loss Lun}}
For unlabeled target samples, we employ a Mean Teacher framework where the teacher is updated as an exponential moving average (EMA) of the student~\cite{tarvainen2017mean}. We minimize the consistency loss $\mathcal{L}_{\text{con}}$ between the student and teacher predictions on $x_u$ under strong ($\zeta'$) and weak ($\zeta$) perturbations, respectively, to enforce perturbation-invariant predictions:

\begin{equation}
\mathcal{L}_{\text{con}} = \mathcal{L}_{1}(\hat{y}^{\text{stu}}_{u}, \hat{y}^{\text{tea}}_{u}).
\end{equation}
Spectral density masking (detailed in Sec.~\ref{subsec:component}) is applied only to the student branch to break channel shortcuts and strengthen cross-channel reasoning.

To mitigate unreliable pseudo-supervision, we introduce Physiologically Constrained Alignment $\mathcal{L}_{\text{PCA}}$ (detailed in Sec.~\ref{subsec:componentB}), a reliability-gated mechanism aligning student-teacher representations based on their plausibility within the target clinical manifold.
The overall unsupervised objective is then defined as:
\begin{equation}
\mathcal{L}_{\text{un}} = \mathcal{L}_{\text{con}} + \lambda_{\text{pca}} \cdot \mathcal{L}_{\text{PCA}},
\end{equation}
where $\lambda_{\text{pca}}$ is a trade-off hyperparameter balancing the consistency regularization and the physiological alignment constraint.

%% file: 5experiments.tex
\section{Evaluation}
\label{sec:evaluation}

\subsection{Evaluation Settings}

\subsubsection{Datasets}
To evaluate the efficacy of object-to-human HSI reconstruction,
as well as the downstream clinical utility of the reconstructed spectra,
we conduct experiments on three categories of datasets:
(i) object-centric source-domain datasets, (ii) human facial target-domain datasets, and (iii) two independent medical benchmarks for downstream validation:

\textbf{Object-centric Source-domain Datasets.} 
We independently employ the NTIRE2020~\cite{arad2020ntire} and NTIRE2022~\cite{arad2022ntire} datasets as two source domains.
NTIRE 2020 comprises 510 images, while NTIRE 2022 contains a larger set of 1,000 images.
These datasets capture diverse real-world scenarios, ranging from indoor objects to natural landscapes and urban structures, characterized by rich spatial textures and high spectral fidelity.
Both datasets are released as 31-band visible-spectrum hyperspectral data at a spatial resolution of 482×512.
Following established protocols, we employ the 31 visible spectral bands (400–700~nm) with a 10~nm stride as the source-domain supervision.

\noindent\textbf{Human Facial Target-domain Datasets.}
The \textbf{Hyper--Skin}~\cite{ng2024hyper}, released through the NeurIPS 2023 Datasets and Benchmarks Track, serves as the target domain for cross-domain adaptation.
It consists of 330 hyperspectral facial images collected from 51 human participants, with spectral coverage from 400–700~nm across 31 bands at a spatial resolution of $1024 \times 1024$. 
With diverse ethnicities and poses, the dataset enables skin spectral reconstruction and physiological quantification of melanin and hemoglobin. 
Following standard protocols, we perform three independent subject-disjoint 7:2:1 train/val/test splits and report mean$\pm$std across splits.

\noindent\textbf{Downstream Medical Datasets.}
To assess whether~\N\ preserves physiologically meaningful spectral information beyond facial imaging, 
we further evaluate the reconstructed hyperspectral images on two independent medical benchmarks.
\begin{itemize} [leftmargin=*]
    \item The \textbf{Multidimensional Choledoch dataset}~\cite{zhang2019multidimensional} is a pathology benchmark consisting of 880 microscopy scenes from 174 patients, 
    with paired RGB and hyperspectral images captured across 60 spectral bands (550–1000~nm) using an AOTF-based system.
    To align with the source setting, we resample the overlapping range to 31 bands via linear interpolation.
    Expert-annotated polygon masks identify cancerous regions, enabling quantitative evaluation of HSI-driven segmentation.
    
    \item The \textbf{HeiPorSPECTRAL dataset}~\cite{studier2023heiporspectral} provides hyperspectral images of 20 physiological porcine organs collected from 11 pigs using the medically certified TIVITA Tissue system.
    Each image covers a $480 \times 640$ field of view with 100 spectral bands spanning 500–1000~nm.
    To match the source configuration, we similarly interpolate the 500–700~nm range to 31 bands to match the channel setting.
    Expert-annotated polygon masks delineate representative tissue regions while excluding confounding artifacts. 
\end{itemize}

\begin{table*}[t]
\centering
\caption{
Comparison with SSDA baselines on NTIRE2020$\rightarrow$Hyper-Skin and NTIRE2022$\rightarrow$Hyper-Skin under 1.5\% and 5\% labeled target training set, evaluated by SSIM, SAM, and PSNR. SSIM is reported in \%, SAM is scaled by \texttimes100, and PSNR is in dB.
}
\setlength{\tabcolsep}{4mm}
\resizebox{\textwidth}{!}{%
\begin{tabular}{l|c|ccc|ccc}
\toprule[1pt]
\multirow{2}{*}{Method} & \multirow{2}{*}{Labeled (Ratio[\%])} & \multicolumn{3}{c|}{NTIRE2020 $\rightarrow$ Hyper-Skin} & \multicolumn{3}{c}{NTIRE2022 $\rightarrow$ Hyper-Skin}  \\ 
\cmidrule(lr){3-5} \cmidrule(lr){6-8}
& & SSIM ($\uparrow$) & SAM ($\downarrow$) & PSNR ($\uparrow$) 
  & SSIM ($\uparrow$) & SAM ($\downarrow$) & PSNR ($\uparrow$) \\
\midrule[0.8pt]
\textit{Source-Only} & 0 (0$\%$) &
\textit{59.87$\pm$4.31} & \textit{67.57$\pm$5.52} & \textit{17.86$\pm$1.29} &
\textit{60.12$\pm$4.45} & \textit{66.89$\pm$5.40} & \textit{18.05$\pm$1.33} \\ 
\midrule[0.8pt]
\textit{Fully Supervised} &\multirow{2}{*}{N (100$\%$)} & \textit{92.37$\pm$0.59} & \textit{12.48$\pm$1.13} & \textit{31.26$\pm$0.41} & \textit{92.37$\pm$0.59} & \textit{12.48$\pm$1.13} & \textit{31.26$\pm$0.41} \\
\textit{Target Finetune} &  & \textit{92.89$\pm$0.44} & \textit{11.73$\pm$0.97} & \textit{31.92$\pm$0.38} & \textit{93.14$\pm$0.41} & \textit{11.25$\pm$0.84} & \textit{32.08$\pm$0.35} \\
\midrule[0.8pt]
% --- 1.5% Setting (Standard Deviations Reduced) ---
Mean Teacher~\cite{tarvainen2017mean} & \multirow{9}{*}{3 (1.5\%)} & 85.30$\pm$2.69 & 21.05$\pm$2.12 & 23.23$\pm$1.17 & 85.65$\pm$2.46 & 20.84$\pm$1.90 & 27.95$\pm$1.29 \\
SADT~\cite{wang2023semi} &  & 81.33$\pm$3.08 & 19.23$\pm$2.06& 22.82$\pm$1.28 & 81.70$\pm$2.96 & 19.67$\pm$1.87& 21.95$\pm$1.20 \\
ADA-SDA~\cite{berthelot2021adamatch}&  & 84.16$\pm$2.51 & 17.56$\pm$1.80 & 24.23$\pm$1.25 & 84.49$\pm$2.39 & 17.12$\pm$1.75 & 26.20$\pm$0.93 \\
SFPM~\cite{ma2023source}&  & 83.69$\pm$2.45 & 19.42$\pm$1.95 & 23.75$\pm$1.36 & 85.05$\pm$2.73 & 18.65$\pm$1.64 & 27.82$\pm$1.06 \\
MCL~\cite{yan2022multi} &  & 84.60$\pm$2.38 & 17.89$\pm$1.73 & 25.20$\pm$1.03 & 84.92$\pm$2.35 & 17.45$\pm$1.56 & 24.78$\pm$1.15 \\
IDM~\cite{li2024inter} &  & 85.35$\pm$2.98 & 16.94$\pm$2.17 & 26.15$\pm$1.41 & 86.36$\pm$2.88 & 16.88$\pm$2.05 & 30.08$\pm$1.14\\
Unmix-SDA~\cite{baghbaderani2023unsupervised} & & 85.82$\pm$2.16 & 16.12$\pm$1.68 & 27.25$\pm$0.97 & 86.41$\pm$2.09 & 16.93$\pm$1.41 & 29.18$\pm$1.04 \\
ProML~\cite{huang2023semi} &  & 87.23$\pm$1.89 & 15.45$\pm$1.57 & 28.22$\pm$0.85 & 87.35$\pm$1.79 & 16.82$\pm$1.37 & 28.95$\pm$0.78 \\
\textbf{\N (Ours)} & & \textbf{88.24$\pm$1.12} & \textbf{14.28$\pm$1.26} & \textbf{28.78$\pm$0.67} & \textbf{89.55$\pm$1.06} & \textbf{13.76$\pm$1.17} & \textbf{30.72$\pm$0.56} \\ 
\midrule[0.5pt]

% --- 5% Setting (Standard Deviations Further Reduced) ---
Mean Teacher~\cite{tarvainen2017mean} & \multirow{9}{*}{10 (5\%)} & 87.94$\pm$1.93 & 18.65$\pm$1.85 & 25.12$\pm$0.98 & 87.82$\pm$1.81 & 18.34$\pm$1.64 & 28.34$\pm$1.03 \\
SADT~\cite{wang2023semi} &  & 84.13$\pm$2.55 & 17.86$\pm$1.81 & 24.38$\pm$1.04 & 84.87$\pm$2.32 & 17.95$\pm$1.58 & 23.96$\pm$1.15 \\
ADA-SDA~\cite{berthelot2021adamatch}&  & 87.17$\pm$1.82 & 16.14$\pm$1.48 & 26.54$\pm$0.89 & 87.63$\pm$1.74 & 15.87$\pm$1.35 & 27.65$\pm$0.81 \\
SFPM~\cite{ma2023source}&  & 86.76$\pm$2.04 & 17.58$\pm$1.52 & 25.82$\pm$1.13 & 87.51$\pm$1.96 & 17.21$\pm$1.49 & 28.27$\pm$0.95 \\
MCL~\cite{yan2022multi} &  & 87.62$\pm$1.78 & 16.32$\pm$1.34 & 26.87$\pm$0.78 & 88.04$\pm$1.62 & 15.96$\pm$1.26 & 26.34$\pm$0.87 \\
IDM~\cite{li2024inter} &  & 88.28$\pm$2.17 & 15.04$\pm$1.93 & 28.14$\pm$1.15 & 89.13$\pm$2.01 & 14.12$\pm$1.84 & 30.65$\pm$1.06\\
Unmix-SDA~\cite{baghbaderani2023unsupervised} & & 88.64$\pm$1.61 & 13.95$\pm$1.39 & 28.63$\pm$0.74 & 89.18$\pm$1.53 & 14.86$\pm$1.31 & 29.87$\pm$0.68 \\
ProML~\cite{huang2023semi} &  & 89.37$\pm$1.42 & 14.12$\pm$1.26 & 29.21$\pm$0.63 & 89.89$\pm$1.25 & 14.75$\pm$1.08 & 29.54$\pm$0.59 \\
\textbf{\N (Ours)} & & \textbf{90.76$\pm$0.92} & \textbf{12.43$\pm$1.05} & \textbf{30.04$\pm$0.51} & \textbf{91.42$\pm$0.87} & \textbf{12.18$\pm$0.98} & \textbf{30.87$\pm$0.48} \\

\bottomrule[1pt]
\end{tabular}
}
\label{ntire_comparison}
\end{table*} 

\subsubsection{Metrics}
To comprehensively evaluate~\N, we conduct quantitative assessments on both hyperspectral reconstruction and downstream semantic segmentation.
For reconstruction, we adopt three standard metrics: Structural Similarity Index (SSIM)~\cite{wang2004image}, Spectral Angle Mapper (SAM)~\cite{kruse1993spectral}, and Peak Signal-to-Noise Ratio (PSNR)~\cite{hore2010image}, which collectively evaluate structural fidelity, spectral accuracy, and overall reconstruction quality.
For segmentation, we use Pixel Accuracy (PA), Mean Pixel Accuracy (MPA), and Mean Intersection over Union (mIoU)~\cite{long2015fully,everingham2010pascal} to evaluate pixel-wise and class-wise classification performance.

\subsubsection{Baselines}
We compare~\N\ with SOTA SSDA methods covering three adaptation paradigms:
(1) pseudo-label consistency methods, including ADA-SDA~\cite{berthelot2021adamatch} and MCL~\cite{yan2022multi};
(2) mixup-based approaches, including IDM~\cite{li2024inter} and SFPM~\cite{ma2023source};
and (3) prototype-driven methods, including Unmix-SDA~\cite{baghbaderani2023unsupervised} and ProML~\cite{huang2023semi}.
To ensure a fair comparison, all baselines are implemented in our reconstruction framework using the same backbone (MST++), identical training iterations (10k), and the same data augmentation and cropping strategy. 
We further implement their consistency and mixup operations on hyperspectral cubes (i.e., band-wise in the spectral domain) and adapt prototype-based baselines to operate in the same intermediate feature space for fair comparison.

\subsubsection{Implementation Details}
All experiments are implemented in PyTorch 1.12.1 under Python 3.7.16 and conducted on an NVIDIA A800 GPU.
We adopt MST++~\cite{cai2022mst++} as the reconstruction backbone.
The network is trained using the Adam optimizer for 10,000 iterations with an initial learning rate of $1 \times 10^{-4}$, decayed via a cosine annealing schedule.
The batch sizes are set to 8 for source-domain data, 8 for unlabeled target data, and 2 for labeled target data.
The momentum for the EMA teacher model is set to $m_{\text{ema}} = 0.99$, and the momentum for updating the prototype bank is set to $m_{\text{pro}} = 0.9$.
We set the softmax temperature to $\tau_p=0.2$ and the gate temperature to $\tau_g=0.1$.
The unsupervised loss weight $\lambda_{\mathrm{un}}(t)$ is ramped up to $\lambda_{\max}{=}0.4$, and we set $\lambda_{\mathrm{pca}}{=}0.3$.
For PCR masking, we set the channel-adaptive masking ratio bounds to $r_{\min}{=}0.3$ and $r_{\max}{=}0.7$.
The prototype banks are constructed with $K{=}16$ target prototypes and $M{=}64$ source prototypes, determined based on ablation studies.
To mitigate overfitting and enhance the utilization of unlabeled data, we apply weak/strong noise perturbations, flipping, and rotation.
The input consists of cropped $256 \times 256$ patches with a stride of 128.
A sliding-window strategy is used to generate the final full-resolution predictions.

\subsection{Overall Performance}
\label{sec:rq1}

Table~\ref{ntire_comparison} reports the quantitative comparison between~\N\ and state-of-the-art baselines.
The Source-Only model performs poorly across all metrics, confirming a severe domain gap between object-centric scenes and human facial tissue.
While existing SSDA methods improve reconstruction quality under limited supervision, ~\N\ consistently achieves the best or second-best performance at both 1.5\% and 5\% labeled target settings.
Notably, with only 1.5\% labeled target data,~\N\ significantly outperforms all SSDA baselines in both SSIM and SAM while maintaining competitive PSNR, demonstrating strong robustness under scarce supervision.
As the labeled ratio increases to 5\%,~\N\ further enlarges its advantage and approaches the performance of the fully-supervised upper bound, indicating high data efficiency and stable cross-domain generalization.
All improvements of~\N\ over SSDA baselines at the 1.5\% setting are statistically significant (paired $t$-test, $p<0.025$, Cohen's $d>2$), with gains of $+$2.20 SSIM over the strongest baseline ProML~\cite{huang2023semi} and $+$3.90 SSIM over Mean Teacher~\cite{tarvainen2017mean}.

\begin{table}[t!]
\centering
\caption{
Ablation studies under extreme label scarcity (1.5\%).
}
\setlength{\tabcolsep}{2mm}
\resizebox{\columnwidth}{!}{%
\begin{tabular}{l|cc|cc}
\toprule[1pt]
\multirow{2}{*}{Method} & \multicolumn{2}{c|}{NTIRE2020 $\rightarrow$ Hyper-Skin}  & \multicolumn{2}{c}{NTIRE2022 $\rightarrow$ Hyper-Skin}  \\ 
\cmidrule(lr){2-3} \cmidrule(lr){4-5}
& SSIM ($\uparrow$) & SAM ($\downarrow$) 
& SSIM ($\uparrow$) & SAM ($\downarrow$) \\
\midrule[0.8pt]

Base (Mean Teacher~\cite{tarvainen2017mean}) & 85.30$\pm$2.69 & 21.05$\pm$2.12 & 85.65$\pm$2.46 & 20.84$\pm$1.90 \\
\midrule[0.8pt]

Base+Block Masking~\cite{devries2017improved} & 86.17$\pm$2.38 & 20.43$\pm$1.47 & 86.49$\pm$1.93 & 19.96$\pm$2.08 \\ 
Base+Grid Masking~\cite{chen2020gridmask}  & 86.92$\pm$1.84 & 19.87$\pm$2.13 & 87.18$\pm$2.76 & 19.13$\pm$1.79 \\ 
Base+PCR & 87.83$\pm$2.41 & 17.96$\pm$2.63 & 88.07$\pm$1.74 & 17.58$\pm$2.44 \\

\midrule[0.4pt] 
Base+PCA (Reverse Gate) & 82.19$\pm$3.17 & 26.74$\pm$2.91 & 82.93$\pm$2.84 & 27.18$\pm$2.69 \\
Base+PCA (w/o Source) & 88.47$\pm$2.13 & 16.09$\pm$2.08 & 88.92$\pm$2.17 & 14.93$\pm$2.34 \\
Base+PCA (w/o Gate) & 88.98$\pm$1.89 & 15.34$\pm$1.96 & 89.84$\pm$1.92 & 13.86$\pm$1.87 \\
Base+PCA (Full) & 89.34$\pm$1.37 & 14.87$\pm$1.71 & 90.56$\pm$1.58 & 13.12$\pm$1.41 \\
\midrule[0.4pt] 
\textbf{Base+PCR+PCA (\N)} & \textbf{90.76$\pm$0.92} & \textbf{12.43$\pm$1.05}  & \textbf{91.42$\pm$0.87} & \textbf{12.18$\pm$0.98} \\

\bottomrule[1pt]
\end{tabular}%
}
\label{ablation}
\end{table}

\subsection{Key Design Assessment}
\label{sec:rq2}
We assess the contribution of key design choices in~\N\ via ablation studies under 1.5\% and 5\% labeled target settings on NTIRE2020-to-Hyper-Skin and NTIRE2022-to-Hyper-Skin adaptation (Table~\ref{ablation}).
Two groups of controlled variants are evaluated: 
(1) mechanism-level variants that replace the channel-adaptive PCR with standard uniform masking strategies (block masking~\cite{devries2017improved} and grid masking~\cite{chen2020gridmask});
(2) component-level variants that ablate key elements of PCA, including source prototypes, the gating mechanism, and gating direction.

\noindent\textbf{Mechanism-level Analysis of Masking.}
To isolate the efficacy of the channel-adaptive masking design, we compare \textit{\ComponentA\ (PCR)} with uniform block and grid masking. 
Uniform strategies provide only marginal improvements, as they ignore channel-wise spectral heterogeneity. 
In contrast, PCR consistently achieves superior performance across all settings, demonstrating that adapting masking ratios to spectral complexity is critical for preserving informative channels and enabling robust spectral reasoning.

\noindent\textbf{Component-level Analysis of PCA.} 
We further validate the internal design of \textit{\ComponentB\ (PCA)} through component-wise ablation.
Reversing the gating direction leads to severe performance degradation, significantly underperforming the baseline, indicating that incorrect gating injects destructive noise rather than effective calibration.
Removing source prototypes (w/o Source) or disabling gating (w/o Gate) also results in clear performance drops.
These findings confirm that both source prototypes and correctly directed gating are essential for stable cross-domain alignment, with the full PCA configuration achieving the best reconstruction fidelity.

\begin{table}[t!]
\centering
\caption{Physiological grounding of PCA prototypes. Inter-prototype variance along Melanin (MI) and Hemoglobin (HI) Indices far exceeds random clustering.}
\label{tab:mihi}
\setlength{\tabcolsep}{3mm}
\begin{tabular}{c|cccc}
\toprule[1pt]
\textbf{Index} & \textbf{Inter-Var} & \textbf{Random-Var} & \textbf{Ratio} & $F$ \\
\midrule[0.8pt]
MI & 1848.30 & 16.51 & $112\times$ & 18.91 \\
HI & 48.49 & 0.39 & $125\times$ & 5.63 \\
\bottomrule[1pt]
\end{tabular}
\end{table}

\noindent\textbf{Physiological Grounding of Learned Prototypes.}
To verify that PCA prototypes encode physiologically meaningful structure rather than statistical clusters, we compute two bio-optical indicators on the 100 nearest tissue pixels per prototype: the Melanin Index $\mathrm{MI}{=}100\!\cdot\!\log_{10}(1/R_{650})$ and the Hemoglobin Index $\mathrm{HI}{=}100\!\cdot\!\log_{10}(R_{600}/R_{540})$~\cite{dawson1980theoretical}. As shown in Table~\ref{tab:mihi}, the inter-prototype variance exceeds random clustering $100\times$ along both axes ($F{=}18.91$ for MI, $F{=}5.63$ for HI; $p{<}0.001$), confirming that prototypes spontaneously organize along melanin--hemoglobin axes—the dominant chromophores of human skin without explicit optical supervision.

\begin{table}[t!]
\centering
\caption{\textbf{Hyper-parameter sensitivity analysis on NTIRE2022$\rightarrow$Hyper-Skin.}
Results are reported on the 1.5\% labeled setting. }
\label{hypers}
\renewcommand{\arraystretch}{1.15}
\resizebox{\columnwidth}{!}{%
\begin{tabular}{l|l|ccc}
\toprule[1pt]
\textbf{Category} & \textbf{Settings} & SSIM$\uparrow$ & SAM$\downarrow$ & PSNR ($\uparrow$)\\

\midrule

\multirow{6}{*}{\textbf{(A) Loss weights}}
& $\lambda_{\mathrm{max}}{=}0.2$ & 89.38 & 13.88 & 30.61 \\
& \textbf{\boldmath $\lambda_{\mathrm{max}}{=}0.4$} & \textbf{89.55} & \textbf{13.76} & \textbf{30.72} \\
& $\lambda_{\mathrm{max}}{=}0.6$ & 89.42 & 13.84 & 30.65 \\
\cline{2-5}
& $\lambda_{\mathrm{pca}}{=}0.1$ & 89.15 & 14.05 & 30.45 \\
& \textbf{\boldmath $\lambda_{\mathrm{pca}}{=}0.3$} & \textbf{89.55} & \textbf{13.76} & \textbf{30.72} \\
& $\lambda_{\mathrm{pca}}{=}0.5$ & 89.32 & 13.92 & 30.58 \\
\midrule

\multirow{4}{*}{\textbf{(B) PCA gating }}
& $\tau_g{=}0.05$ & 87.85 & 15.12 & 29.45 \\
& \textbf{\boldmath $\tau_g{=}0.1$} & \textbf{89.55} & \textbf{13.76} & \textbf{30.72} \\
& $\tau_g{=}0.2$ & 88.10 & 14.95 & 29.65 \\
& $\tau_g{=}0.5$ & 86.45 & 16.80 & 28.20 \\
\midrule

\multirow{7}{*}{\textbf{(C) Prototype bank size}}
& $K{=}8$, $M{=}64$ & 88.42 & 14.65 & 29.95 \\
& \textbf{\boldmath $K{=}16$, $M{=}64$} & \textbf{89.55} & \textbf{13.76} & \textbf{30.72} \\
& $K{=}32$, $M{=}64$ & 88.92 & 14.25 & 30.35 \\
& $K{=}64$, $M{=}64$ & 88.15 & 14.88 & 29.65 \\
\cline{2-5}
& $K{=}16$, $M{=}32$ & 87.45 & 15.32 & 29.10 \\
& \textbf{\boldmath $K{=}16$, $M{=}64$} & \textbf{89.55} & \textbf{13.76} & \textbf{30.72} \\
& $K{=}16$, $M{=}128$ & 89.10 & 14.05 & 30.45 \\
\midrule
\multirow{4}{*}{\textbf{(D) Prototype init.}}
& \textbf{K-means} & \textbf{89.55} & \textbf{13.76} & \textbf{30.72} \\
& Mean+noise   & 89.21 & 14.13 & 30.48 \\
& Random       & 89.08 & 14.25 & 30.42 \\
& Source-biased & 87.67 & 16.63 & 28.91 \\
\bottomrule[1pt]
\end{tabular}
}
\end{table}

{
{
\subsection{Sensitivity Analysis}
Table~\ref{hypers} evaluates the robustness of~\N\ with respect to key hyperparameters, including loss weights ($\lambda_{\mathrm{max}}, \lambda_{\mathrm{pca}}$), the PCA gating temperature ($\tau_g$), prototype bank configurations ($K, M$), and prototype initialization.

\noindent\textbf{Loss Weights ($\lambda_{\max}, \lambda_{\mathrm{pca}}$).}
Varying $\lambda_{\max}$ and $\lambda_{\mathrm{pca}}$ leads to only minor changes in performance, with the best balance at $\lambda_{\max}{=}0.4$ and $\lambda_{\mathrm{pca}}{=}0.3$, indicating a robust trade-off between supervised reconstruction and unsupervised regularization.

\noindent\textbf{PCA Gating Temperature ($\tau_g$).}
A moderate value ($\tau_g{=}0.1$) yields the highest SSIM (89.55\%), whereas a larger temperature ($\tau_g{=}0.5$) causes a sharp drop (86.45\%), indicating that an overly smooth gate weakens the contrast between target-consistent and source-like locations and degrades prototype alignment.

\noindent\textbf{Prototype Bank Size ($K, M$).}
The configuration with $K{=}16$ and $M{=}64$ performs best, whereas reducing the source bank ($M{=}32$) weakens source-reference coverage and excessively increasing the target bank ($K{=}64$) introduces redundancy and unstable assignments, indicating that target diversity and source coverage must be jointly balanced.

}

\begin{table}[t]
\centering
\caption{
Downstream reconstruction under 10\% supervision, comparing SSDA baselines and the fully supervised upper bound.
}
\setlength{\tabcolsep}{5mm}
\resizebox{\columnwidth}{!}{%
\begin{tabular}{l|cc|cc}
\toprule[1pt]
\multirow{2}{*}{Method} & \multicolumn{2}{c|}{NTIRE2022 $\rightarrow$ Choledoch} & \multicolumn{2}{c}{NTIRE2022 $\rightarrow$ HeiPorSPECTRAL} \\
\cmidrule(lr){2-3} \cmidrule(lr){4-5}
& SSIM ($\uparrow$) & SAM ($\downarrow$) 
& SSIM ($\uparrow$) & SAM ($\downarrow$) \\
\midrule[0.7pt]
Baseline~\cite{tarvainen2017mean}   & 78.17 $\pm$ 2.89 & 18.23 $\pm$ 1.34 & 88.34 $\pm$ 2.56 & 14.63 $\pm$ 1.18 \\ 
\midrule[0.7pt]
MCL~\cite{yan2022multi}             & 79.93 $\pm$ 2.67 & 16.37 $\pm$ 1.12 & 90.71 $\pm$ 2.23 & 12.91 $\pm$ 1.04 \\
IDM~\cite{li2024inter}              & 80.64 $\pm$ 2.41 & 15.82 $\pm$ 0.96 & 91.28 $\pm$ 1.98 & 12.08 $\pm$ 0.93 \\
Unmix-SDA~\cite{baghbaderani2023unsupervised} & 81.28 $\pm$ 2.19 & 14.94 $\pm$ 0.88 & 93.03 $\pm$ 1.74 & 11.42 $\pm$ 0.87 \\
ProML~\cite{huang2023semi}          & 82.41 $\pm$ 1.93 & 13.76 $\pm$ 0.79 & 93.17 $\pm$ 1.53 & 10.56 $\pm$ 0.71 \\
\textbf{\N (Ours)}                  & \textbf{83.89 $\pm$ 1.47} & \textbf{12.13 $\pm$ 0.64} & \textbf{94.12 $\pm$ 1.09} & \textbf{9.18 $\pm$ 0.56} \\
\midrule[0.8pt]
Fully Supervised                    & 85.34 $\pm$ 1.12 & 11.08 $\pm$ 0.43 & 95.83 $\pm$ 0.76 & 8.43 $\pm$ 0.39 \\ 
\bottomrule[1pt]
\end{tabular}
}
\label{medical_reconstruction}
\end{table}

\subsection{Downstream Validation}
\label{sec:rq4}

\begin{figure*}
\centering
\includegraphics[width=1\textwidth]{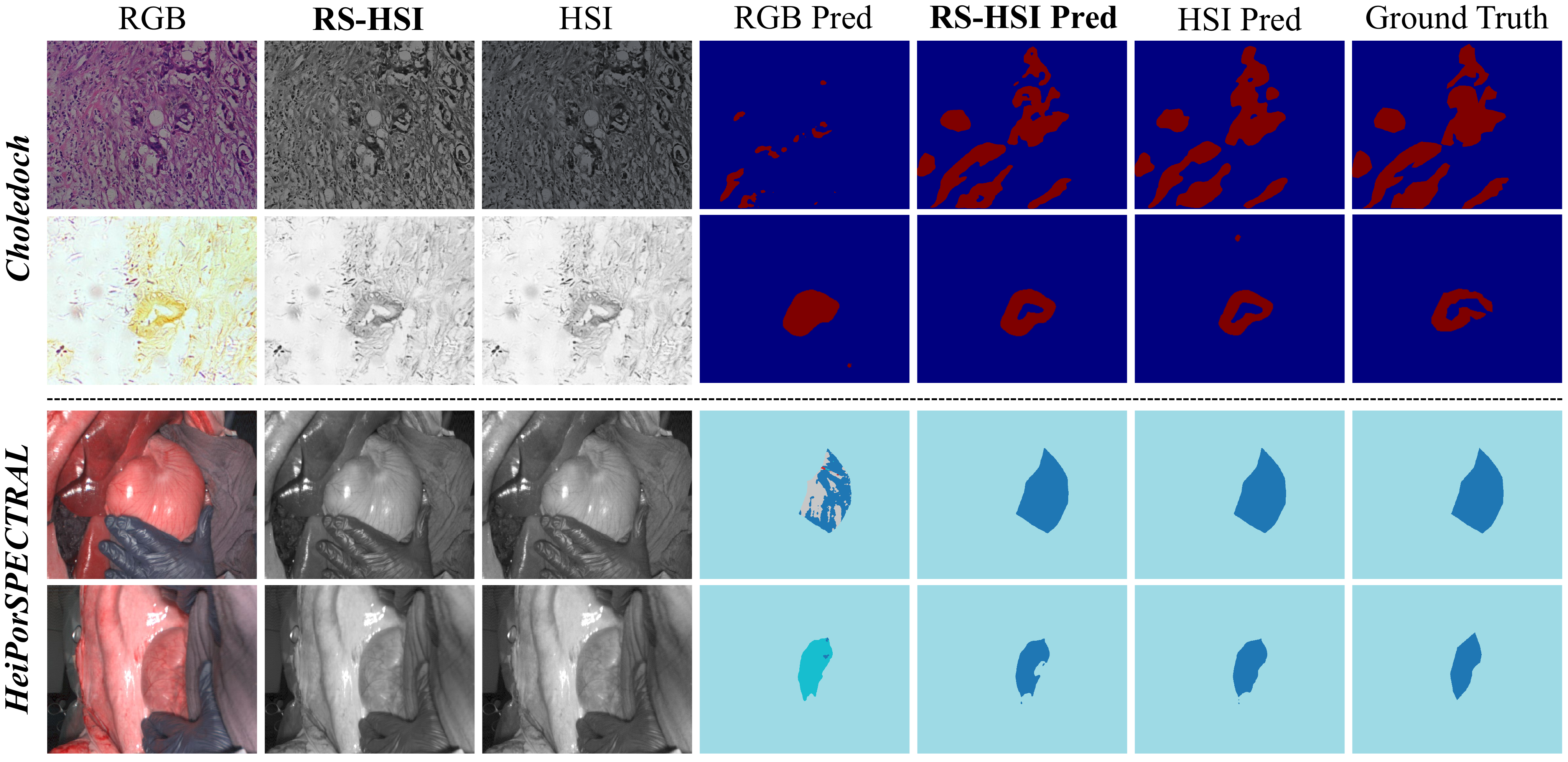}
\caption{Qualitative comparison of segmentation results on Choledoch and HeiPorSPECTRAL using three input modalities: 
RGB, reconstructed HSI (RS-HSI), and raw HSI.
RS-HSI yields sharper boundaries and more spatially consistent predictions than RGB, matching the performance of raw HSI.
}
\Description{Qualitative segmentation results comparing RGB, reconstructed HSI, and raw HSI inputs against ground truth on two medical datasets.}
\label{downstreamvis}
\end{figure*}

The advantages of spectral information are particularly evident in challenging medical scenarios where RGB fails to capture subtle tissue differences.
On Choledoch~\cite{zhang2019multidimensional}, which involves fine-grained glandular structures with low visual contrast, RGB-based segmentation often suffers from boundary ambiguity and under-segmentation.
On HeiPorSPECTRAL~\cite{studier2023heiporspectral}, intraoperative conditions such as illumination variation and tissue deformation further degrade RGB performance.
These challenges highlight the limitation of RGB in distinguishing physiologically similar tissues and motivate evaluating whether reconstructed HSI can recover clinically useful spectral cues.

\noindent\textbf{Evaluation Pipeline.}
We conduct downstream semantic segmentation experiments on three input modalities: RGB, raw HSI, and reconstructed HSI (RS-HSI).
For fair comparison, all modalities are fed into the same segmentation architecture under identical training and evaluation protocols.
We adopt a standard UNet~\cite{ronneberger2015u} as the backbone and only modify the first convolutional layer to accommodate different input dimensionalities.

\noindent\textbf{HSI Reconstruction from RGB.}
We apply multiple SSDA methods, including~\N, to reconstruct hyperspectral representations from RGB under the standard 10\% labeled setting~\cite{cheplygina2019not,jiao2024learning}.
As reported in Table~\ref{medical_reconstruction},~\N\ consistently achieves the highest SSIM and PSNR on both Choledoch and HeiPorSPECTRAL, approaching the fully supervised upper bound.
These results verify that~\N\ recovers both structural fidelity and spectral accuracy under limited supervision, providing a reliable basis for downstream evaluation.

\begin{table}[t]
\centering
\caption{Downstream segmentation performance using different input modalities on Choledoch and HeiPorSPECTRAL.
}
\label{downstreamtabble}
\resizebox{\columnwidth}{!}{
\begin{tabular}{l|l|ccc}
\toprule
\textbf{Dataset} & \textbf{Input} 
& \textbf{PA ($\uparrow$)} 
& \textbf{MPA ($\uparrow$)} 
& \textbf{mIoU ($\uparrow$)} \\
\midrule
\multirow{3}{*}{Choledoch~\cite{zhang2019multidimensional}} 
& RGB     & 87.9$\pm$1.0 & 83.5$\pm$1.5 & 75.6$\pm$1.2 \\
& \textbf{RS-HSI} & \textbf{90.5$\pm$0.8} & \textbf{86.0$\pm$1.2} & \textbf{79.3$\pm$1.0}\\
& HSI     & 92.3$\pm$0.7 & 88.1$\pm$1.1 & 81.7$\pm$1.0 \\

\midrule
\multirow{3}{*}{HeiPorSPECTRAL~\cite{studier2023heiporspectral}} 
& RGB     & 82.3$\pm$1.3 & 66.3$\pm$2.0 & 56.5$\pm$1.6 \\
& \textbf{RS-HSI} & \textbf{84.6$\pm$1.1} & \textbf{68.7$\pm$1.8} & \textbf{59.9$\pm$1.4}\\
& HSI     & 86.7$\pm$1.0 & 70.2$\pm$1.6 & 61.7$\pm$1.3 \\

\bottomrule
\end{tabular}
}
\end{table}

\noindent\textbf{Segmentation Performance.}
We further evaluate semantic segmentation using RGB, raw HSI, and RS-HSI as inputs on both datasets.
As shown in Table~\ref{downstreamtabble} and Figure~\ref{downstreamvis}, RS-HSI significantly outperforms RGB, boosting mIoU by 3.7 and 3.4 on Choledoch and HeiPorSPECTRAL, respectively, while paralleling raw HSI performance. These results confirm that~\N\ effectively bridges the spectral gap, enabling reliable clinical interpretation where direct HSI acquisition is impractical.

\subsection{Robustness to Metameric Ambiguity}
\label{sec:metameric}
To directly probe challenge C2 (Sec.~\ref{sec:intro}), where spectrally distinct tissues map to near-identical RGB observations, we stress-test \N\ on 92 \textit{metameric hard pairs} curated from NTIRE2022-to-Hyper-Skin adaptation.
These pairs have near-indistinguishable RGB inputs (RGB $L_2{\le}0.0017$) yet substantially different ground-truth spectra (SAM${\ge}8.91^\circ$).
As reported in Table~\ref{tab:metameric}, \N\ attains the lowest reconstruction SAM on this subset ($8.52^\circ$), ${\sim}10\%$ below Mean Teacher~\cite{tarvainen2017mean} and ProML~\cite{huang2023semi}, a statistically significant margin (paired $t$-test, $p{=}1.91{\times}10^{-10}$).
Together with the reverse-gate ablation (Table~\ref{ablation}), this confirms that the reliability gate in PCA is the key mechanism for resolving metameric ambiguity.

\begin{table}[t!]
\centering
\caption{Reconstruction SAM on the 92 metameric hard pairs. }
\label{tab:metameric}
\resizebox{\columnwidth}{!}{%
\begin{tabular}{l|ccc}
\toprule[1pt]
Method & Mean Teacher~\cite{tarvainen2017mean} & ProML~\cite{huang2023semi} & \textbf{\N (Ours)} \\
\midrule[0.8pt]
SAM ($^\circ$, $\downarrow$) & 9.52 & 9.61 & \textbf{8.52} \\
\bottomrule[1pt]
\end{tabular}%
}
\end{table}

%% file: 7conclusion.tex
\section{Conclusion}
\label{sec:conclusion}

In this study, we reposition object-to-human hyperspectral reconstruction from a reflectance alignment problem to one of physiology-aware representation learning.
Building on this perspective, we design \N, a physiology-aware hyperspectral reconstruction paradigm structured around two pillars:
\textit{\ComponentA}, which reconstructs cross-channel semantics by disrupting reflectance-driven shortcuts and enforcing physiologically consistent inter-channel inference;
and \textit{\ComponentB}, which constrains inherently ambiguous reconstruction by suppressing physiologically invalid yet RGB-consistent explanations and steering predictions toward clinically grounded spectra.
Across NTIRE2020$\rightarrow$Hyper-Skin and NTIRE2022$\rightarrow$Hyper-Skin, \N\ outperforms state-of-the-art baselines, achieving up to $+2.20$ SSIM and $-3.06$ SAM
with only 1.5\% labeled supervision.
As a result, \N\ bridges the object-human domain gap by enforcing physiological constraints, revealing that reliable object-to-human transfer arises from excluding physiologically implausible channel interpretations rather than surface appearance alignment.